%% file: main.tex
\pdfoutput=1
\documentclass[10pt,twocolumn,letterpaper]{article}

\usepackage{cvpr}              

\usepackage{svg}
\usepackage{graphicx}
\usepackage{amsmath}
\usepackage{amssymb}
\usepackage{booktabs}
\usepackage{xcolor}
\usepackage{siunitx}
\usepackage{epsfig}
\usepackage{multirow}
\usepackage{multicol}
\usepackage{makecell}
\usepackage{hyphenat}
\usepackage{tabularray}

\definecolor{cvprblue}{rgb}{0.21,0.49,0.74}

\usepackage[pagebackref,breaklinks,colorlinks,citecolor=cvprblue]{hyperref}

\def\paperID{4332} 
\def\confName{CVPR}
\def\confYear{2024}

\title{TAB: Text-Align Anomaly Backbone Model for Industrial Inspection Tasks}

\author{Ho-Weng Lee\textsuperscript{1} \quad Shang-Hong Lai\textsuperscript{2}\vspace{0.3em} \\
{Department of Computer Science, National Tsing Hua University, Taiwan}\vspace{0.3em}\\
{\tt\small howeng0901@gapp.nthu.edu.tw, lai@cs.nthu.edu.tw}\\
}

\begin{document}
\maketitle
\input{sec/abstract}
\input{sec/introduction}
\input{sec/relatedwork}
\input{sec/method}
\input{sec/expriments}

\input{sec/conclusion}

{
    \small
    \bibliographystyle{ieeenat_fullname}
    \bibliography{main}
}


\end{document}

%% file: sec/abstract.tex
\begin{abstract}
In recent years, the focus on anomaly detection and localization in industrial inspection tasks has intensified. While existing studies have demonstrated impressive outcomes, they often rely heavily on extensive training datasets or robust features extracted from pre-trained models trained on diverse datasets like ImageNet. In this work, we propose a novel framework leveraging the visual-linguistic CLIP model to adeptly train a backbone model tailored to the manufacturing domain. Our approach concurrently considers visual and text-aligned embedding spaces for normal and abnormal conditions. The resulting pre-trained backbone markedly enhances performance in industrial downstream tasks, particularly in anomaly detection and localization. Notably, this improvement is substantiated through experiments conducted on multiple datasets such as MVTecAD, BTAD, and KSDD2. Furthermore, using our pre-trained backbone weights allows previous works to achieve superior performance in few-shot scenarios with less training data. The proposed anomaly backbone provides a foundation model for more precise anomaly detection and localization.
\end{abstract}

%% file: sec/introduction.tex
\section{Introduction}
\label{sec:intro}

\begin{figure}[ht]
  \includegraphics[scale=0.23]{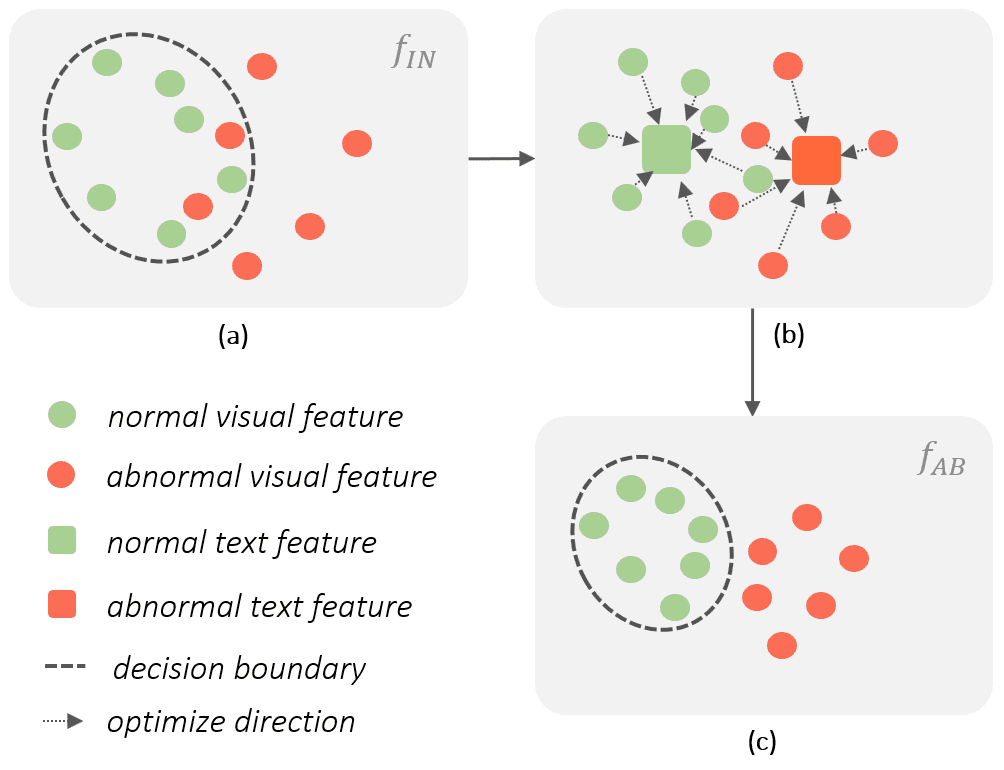}   
   \caption{Overview of our Image visual features and CLIP text features in latent space. (a) ImageNet pre-trained initial visual features, (b) add CLIP text features and optimize similarity of text and visual features, and (c) our Text-align Anomaly Backbone(TAB) without text encoder in downstream tasks.}
   \label{fig:optimized_feature}
\end{figure}

In industrial inspection, three crucial research topics in computer vision are anomaly detection, anomaly localization, and defect classification with applications in various fields. Anomaly Detection (AD) aims to classify whether an image is normal or abnormal. It poses a long-tail problem since it is challenging to collect anomalous samples, and the types of anomalies can vary from subtle changes, such as thin scratches, to more significant structural defects. As a result, most prior works focus on learning the nominal distribution from available standard samples and treating features as abnormal if their deviation from the normal distribution is significant enough. On the other hand, Anomaly Localization (AL) aims to segment and identify anomalous regions at the pixel level. Defect Classification (DC) is another critical task that categorizes the identified anomalies into specific classes based on their characteristics. These three tasks often work together in a pipeline to provide a comprehensive solution for industrial inspection tasks. Collectively, they contribute to improving the efficiency and accuracy of automated inspection systems.


\begin{figure*}[t]
\begin{center}
\includegraphics[scale=0.095]{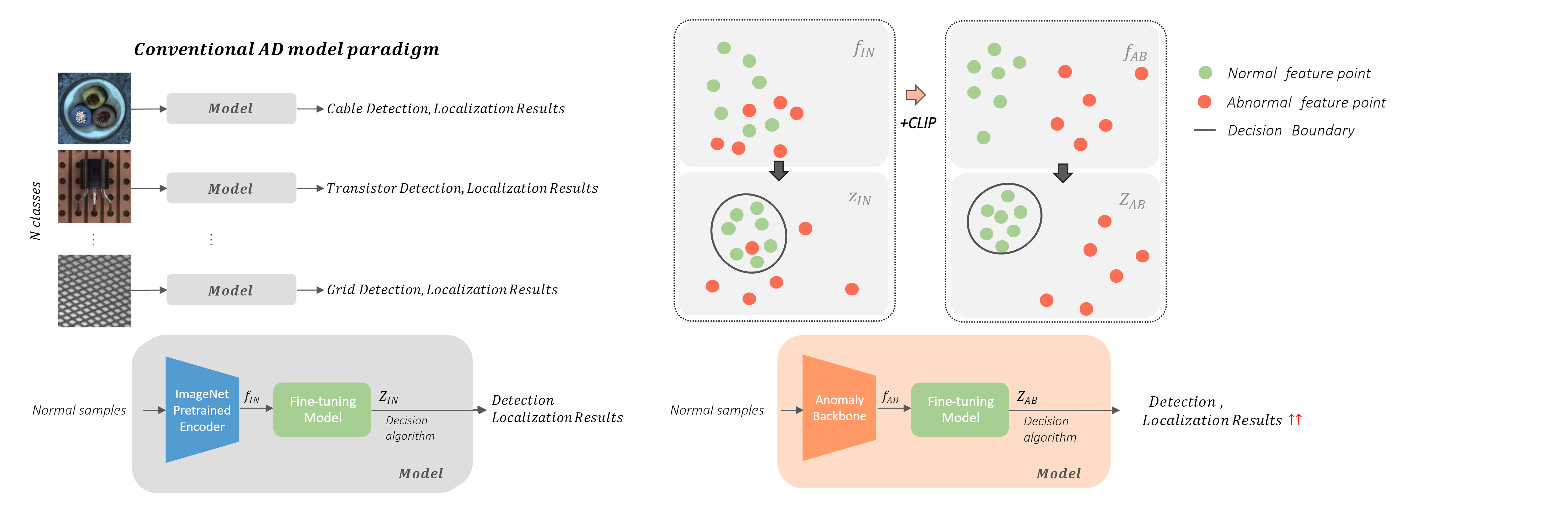}
\end{center}
   \caption{The figure provides a comparative overview of the conventional Anomaly Detection/Anomaly Localization (AD/AL) paradigm and our proposed Text-align Anomaly Backbone (TAB). The left block illustrates that prior works require N models for N classes, achieved by fine-tuning the ImageNet pre-trained features, denoted as ${f_{IN}}$, through a feature extractor into ${Z_{IN}}$. However, these models struggle to distinguish anomalous features from normal ones. The right bottom block demonstrates how TAB improves upon this by generating better-distributed features ${f_{AB}}$ and fine-tuned features ${Z_{AB}}$, outperforming those obtained with the ImageNet pre-trained model.}

\label{fig:structure}
\end{figure*}

Cutting-edge anomaly detection and localization methods, such as those \cite{jeong2023winclip, defard2021padim, roth2022towards, liu2023simplenet, deng2022anomaly}, leverage features extracted from a pre-trained ImageNet \cite{deng2009imagenet} backbone. Some even construct an additional model atop the backbone to re-project the features into a novel latent space. This strategy enhances the model’s ability to delineate normal and abnormal features, thereby improving anomaly detection. Anomaly localization is achieved by creating an anomaly map that matches the input image’s dimensions. This map identifies anomalous regions and selects the highest score as the anomaly detection result for the input image. However, these research trends increase the model’s parameter count and inference cost, rendering them impractical for real-world industrial production lines.

Numerous studies, such as \cite{roth2022towards, deng2022anomaly, defard2021padim}, primarily utilize intermediate feature maps from the backbone to construct their anomaly maps. The deeper layers are often overlooked due to the significant domain gap between the source and target industrial datasets, which arises because the deeper weights of the backbone are trained on natural datasets in a classification context. The majority of previous work relies heavily on the backbone for representation extraction. Therefore, this study aims to address the domain gap between the source domain (ImageNet natural data) and the target domain (industrial data) by focusing on upstream tasks, rather than downstream tasks or backward works.


In this work, we introduce a novel pre-training paradigm, the \textbf{T}ext-Align \textbf{A}nomaly \textbf{B}ackbone (TAB), which employs a robust multi-modal CLIP \cite{radford2021learning} text encoder to align industrial visual and text information. Our approach optimizes the Anomaly Backbone to bring normal visual features closer to their corresponding normal text features in the latent space. Similarly, anomalous visual features are aligned with anomalous text features. This alignment ensures that normal and anomalous feature clusters are well-separated, as depicted in Fig. \ref{fig:optimized_feature}. After pre-training, the CLIP text encoder is detached, retaining only the weights of the visual feature extractor. When state-of-the-art methods utilize our weights in their feature extractors, they can provide a generalized feature representation sensitive to anomalous features, as illustrated in Fig. \ref{fig:structure}. This benefits industrial inspection downstream tasks by improving performance without the need for additional models, data, or fine-tuning. Our weights serve as a plug-and-play component, ensuring rapid adaptability and high generalization in few-shot scenarios. 

The contributions of this paper are listed as follows:
\begin{itemize}
\item We introduce the novel Text-Align Anomaly Backbone (TAB), a pre-training framework that trains a backbone to discriminate out-of-distribution (defect) features in an unsupervised manner. To our knowledge, we are the first to investigate the backbone while considering industrial anomaly information.

\item We propose an Industrial Domain Prompt Association (IDPA) to provide a textual supervision and Anomaly-Text-Aware pre-training strategy that efficiently leverages CLIP to guide and constrain feature representation learning by aligning visual and text features in both normal and abnormal terms.

\item We demonstrate that our proposed work, TAB, when integrated with existing state-of-the-art methods, can deliver superior performance in various industrial inspection tasks. These include anomaly detection, localization, and defect classification on benchmark datasets, such as MVTecAD\cite{bergmann2019mvtec}, BTAD\cite{mishra2021vt}, KSDD2\cite{bovzivc2021mixed}, and MixedWM38\cite{mixedwm38dataset}. Our approach is proved to be effective for few-shot and cross-dataset testing scenarios.
\end{itemize}

%% file: sec/relatedwork.tex
\section{Related Work}
\label{sec:relatedwork}

\subsection{Anomaly Detection Methods}
Anomaly detection problem is cast as an out-of-distribution (OOD) detection problem. OOD aims to approximate a decision boundary as tight as possible around the normal set while excluding unseen samples from other classes or distributions. However, anomalous data is usually inaccessible or insufficient.
Thus, only normal data are available for the training, making the model focused on learning to model normal features. Once an out-of-distribution sample is detected in the feature space, we can treat it as an anomaly.

Most prior works heavily rely on the quality of the extracted features. Embedding-based methods \cite{defard2021padim} employed the patch features of pre-trained backbone and multivariate Gaussian distributions to estimate the feature distribution of nominal data. In the inference stage, the embeddings of irregular patches are assumed to be distributed away from the model for normal samples. For the distillation-based methods \cite{deng2022anomaly, batzner2023efficientad}, the teacher network pre-trained on ImageNet plays a crucial role in the whole setting. \cite{deng2022anomaly} uses a reverse flow that avoids the confusion caused by the same filters and prevents the propagation of anomaly perturbation to the student model, whose structure is similar to reconstruction networks. A sample is determined as normal by checking the consistency in behavior between the teacher and student features. 

Flow-based methods \cite{gudovskiy2022cflow, yu2021fastflow}, such as normalizing flows(NF), are invertible neural networks that learn to transform the high dimensions and complex extracted features into a simpler distribution, e.g., Gaussian distribution, and then use the normal distribution to decide a threshold as the decision boundary of normal and anomalous features. For the memory-based methods \cite{roth2022towards, xie2023pushing}, \cite{roth2022towards} proposed the state-of-the-art algorithm, which builds a memory bank to store the representation of normal features and then compares the feature similarity between the inference sample features with all the stored features to compute the corresponding anomaly scores. No matter which method, a feature representation that well discriminates normal and anomalous samples will benefit the downstream tasks.

\begin{figure*}[t]
\begin{center}
\includegraphics[scale=0.38]{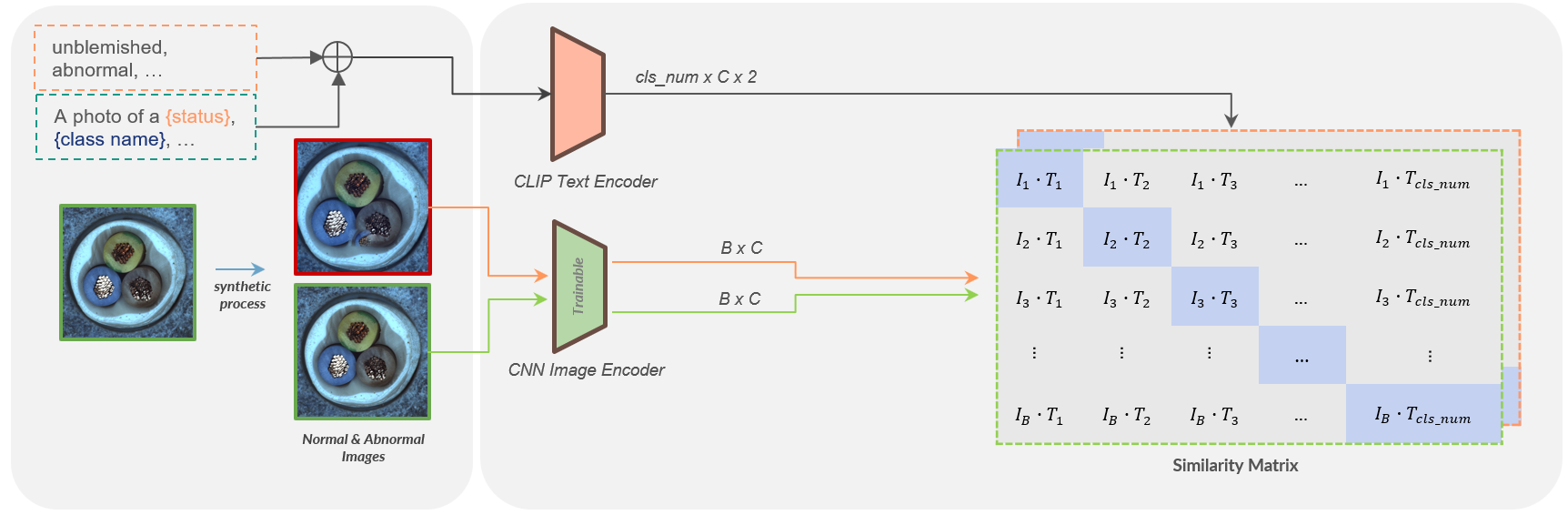}
\end{center}
   \caption{The figure presents the proposed framework of the Text-align Anomaly Backbone (TAB). Our approach incorporates a Convolutional Neural Network (CNN) image encoder and a Contrastive Language–Image Pretraining (CLIP) text encoder. The objective is to align and adjust the visual embedding to match the associated text embedding closely. Our pre-trained CNN image encoder can represent normal and abnormal features by considering the corresponding text embedding. This makes it more robust than conventional models pre-trained on ImageNet.}
\label{fig:TAB_framework}
\end{figure*}


\subsection{Pre-trained on ImageNet}
ImageNet dataset\cite{deng2009imagenet} plays a crucial role in developing and evaluating various computer vision tasks, including object detection, object classification, and semantic segmentation. It comprises over 15 million labeled high-resolution natural images belonging to roughly 22,000 categories. The variety of categories has contributed to modern computer vision models' robustness and generalization capabilities. The pre-trained model brings great benefits for transferring it to specific downstream tasks. Most of the AD prior works mentioned in section 2.1 are based on the feature extractor backbone pretrained on ImageNet. However, various categories in the ImageNet dataset will bring a marginal bias into the downstream domain adaptation and cause an increase in the model scale to overcome the problem. \cite{guo2023mldfr, liu2023simplenet, yang2023anomaly} also mentioned and pointed out that the domain gap between the ImageNet dataset and industrial datasets may limit the model performance. \cite{liu2023simplenet} try to minimize the bias and domain gap by re-project the ImageNet features into a new space by adopting a feature adapter, which is not applicable in another case. 
Thus, this paper aims to address the industrial domain downstream tasks, such as anomaly detection and localization, by proposing a framework that can produce a domain-specific feature extractor backbone to generate a high-quality feature embedding, which is more sensitive to industrial inspection tasks.

\subsection{Foundation Multi-Modal - CLIP}
Contrastive Language–Image Pre-training (CLIP)\cite{radford2021learning} is trained
with massive image and text pairs from the internet with noise contrastive estimation, where image and text pairs from the same sample are used as positive examples, and all the others in the same batch are treated as negatives. Both CLIP image and text representations are projected to a shared embedding space, with the key idea that natural language can be leveraged as a flexible prediction space to help generalize and transfer knowledge, and the continuous conceptual space of images is mapped to the discrete symbolic space of text. As a result, CLIP has shown great success applied to tasks such as zero-shot classification and cross-modal retrieval and has also been extended to cross-modal generation. In this paper, we will leverage the CLIP text encoder to help pre-train our anomaly backbone to align industrial visual features to text features by simultaneously considering normal and abnormal information.

%% file: sec/method.tex
\section{Proposed Method}
\label{sec:method}


In this paper, our primary objective is to develop a pre-trained backbone capable of generating feature representations aptly suited for industrial inspection tasks. An overview of the proposed framework is illustrated in Figure \ref{fig:TAB_framework}. The framework primarily consists of two components: pre-processing and training. The pre-processing component is responsible for generating Synthetic Anomaly Samples (SAS) and Industrial Domain Prompt Association (IDPA). Section \ref{SAS} will provide detailed descriptions of the synthetic flows. Section \ref{IDPA} will elaborate on the industrial prompt ensemble association and samples. The training component, discussed in Section \ref{Loss_OF}, introduces the objective functions of our proposed framework. We will elucidate aligning visual and text features using a similarity matrix. The steps of our Anomaly Backbone pre-training are outlined in the Algorithm flow, which can be found in the supplementary materials.

\subsection{Synthetic Anomaly Samples (SAS)\label{SAS}}

\begin{figure}
\begin{center}
\includegraphics[scale=0.105]{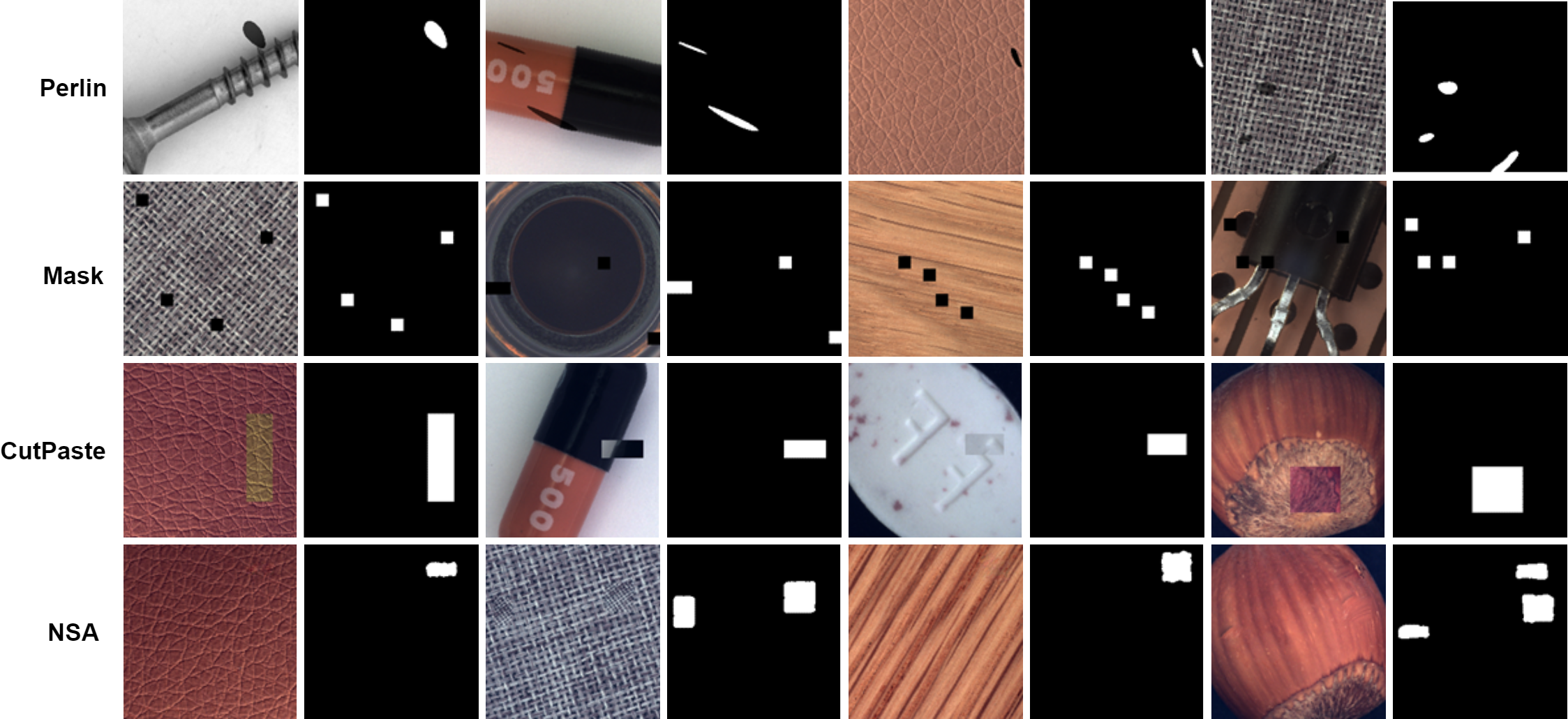}
\end{center}
   \caption{Illustration showcases four distinct types of synthetic image samples, each with their respective pixel-level ground truth. NSA ensures a natural and realistic appearance in these synthetic samples.}
\label{fig:synthetic_samples}
\end{figure}

Our work necessitates the incorporation of anomalous samples. To circumvent the high costs associated with using real-world anomalous samples, we employ an anomaly synthesis mechanism. For the training phase, we consider real normal images, denoted as ${I_n}$. We integrate strategies such as image masking and pseudo-anomaly for the anomalous samples into the anomaly sample synthesis process. Upon reviewing our experimental results (Table \ref{tab:syntheticmethods}) and visualizations of synthetic samples (Figure \ref{fig:synthetic_samples}), we conclude that the Natural Synthetic Anomaly (NSA) method should be included in our pre-training process. This decision was based on the observation that other methods produced unsatisfactory anomalous samples, either due to noticeable artifacts or because the synthesized defects were too subtle compared to real-world defects. In contrast, the NSA method generated anomalies that were more natural in appearance. This approach ensures that our model is trained on diverse anomalies, enhancing its ability to detect and classify anomalies in industrial inspection tasks. 

NSA first randomly selects a source image ${I_{src}}$ which is the same class with ${I_n}$, followed by randomly cropping a patch denoted as ${P_{src}}$, resizing it randomly and pasting it to the input image $I_n$, NSA\cite{schluter2022natural} integrates Poisson image editing\cite{perez2023poisson} to blend scaled patches seamlessly, denoted the pasted images as synthetic abnormal image $I_a$. As mentioned above, this process will be repeated several times to generate various defects on $I_a$. 

After the synthetic process, an input image pair (${I_n, I_a}$) will be fed into CNN image visual encoder $E_V$ to extract a pair of image features. Both image features are in ${\mathbb{R}^{B*C}}$, where $B$ and $C$ are the batch size and channel size of features.

\subsection{Industrial Domain Prompt Association (IDPA)\label{IDPA}}


Inspired by WinCLIP\cite{jeong2023winclip}, we have designed an associative input prompt with the template: “A photo of a \{\textbf{prompt\_state}\} \{\textbf{class\_name}\}.” as \ref{fig:TAB_framework} shown. Here, \{\textbf{class\_name}\} can be replaced by the label name from the training data, such as ‘transistor,’ ‘grid,’ or ‘cable.’ The \{\textbf{prompt\_state}\} is designed to reflect two primary states in a symmetric manner: \{\textbf{prompt\_normal}\} and \{\textbf{prompt\_abnormal}\}, which could be ‘unblemished \{\textbf{class\_name}\}’ and ‘blemished \{\textbf{class\_name}\},’ respectively. More detailed examples of prompts are provided in the supplementary materials.

For the input images, denoted as $X_n$, and their corresponding class names, we apply the Industrial Domain Prompt Association (IDPA) to generate a list of normal and abnormal text sentences, denoted as $T_n$ and $T_a$ in Eq.(\ref{eq:eq1}). This approach ensures a comprehensive and precise representation of normal and abnormal states, enhancing the model’s ability to detect and classify anomalies accurately.

\begin{align}
\begin{split}
    {T_n}&=\{{{T^{ screw}_n}}, {{T^{ cable}_n}}, {{T^{ grid}_n}}, ... \}, 
    \\
    {T_a}&=\{{{T^{ screw}_a}}, {{T^{ cable}_a}}, {{T^{ grid}_a}}, ... \},
    \\    
\end{split}
\label{eq:eq1}
\end{align}

For each class, the normal and abnormal text sentences are described below:
\begin{align}
\begin{split}
    {T^{screw}_n}&=\{{{T^{screw}_{n,1}}}, {{T^{screw}_{n,2}}}, {{T^{screw}_{n,3}}}, ...\},
    \\
    {T^{screw}_a}&=\{{{T^{screw}_{a,1}}}, {{T^{screw}_{a,2}}}, {{T^{screw}_{a,3}}}, ...\},
    \\
\end{split}
\label{eq:eq2}
\end{align}

For each class, the text sentence is fed into the CLIP text encoder, and mean pooling is applied to the extracted features. $f^{T}_n$ and $f^{T}_a$, both are output features of normal and abnormal text sentences from CLIP text encoder $E_T$, both are in the shape of ${\mathbb{R}^{K*C}}$, where $K$ and $C$ refer to the number of different class names and the number of feature channels, respectively.


\definecolor{green}{RGB}{144,238,144}
\definecolor{gray}{RGB}{128,128,128}

\begin{table*}
\centering
\renewcommand{\arraystretch}{0.5}
\tabcolsep=0.13cm
\
\begin{tabular}{c|cc|cc|cc|cc|cc}

\toprule
&\multicolumn{2}{c}{PaDiM\cite{defard2021padim}} &\multicolumn{2}{c}{RD4AD\cite{deng2022anomaly}} &\multicolumn{2}{c}{SimpleNet\cite{liu2023simplenet}} &\multicolumn{2}{c}{RegAD(4shots)\cite{huang2022registration}} &\multicolumn{2}{c}{RegAD(8shots)\cite{huang2022registration}}\\
\cmidrule(lr){2-3} \cmidrule(lr){4-5} \cmidrule(l){6-7} \cmidrule(l){8-9} \cmidrule(l){10-11}
Pre-trained &{ImgN.} &{TAB} &{ImgN.} &{TAB} &{ImgN.} &{TAB} &{ImgN.} &{TAB*} &{ImgN.} &{TAB*}\\
\midrule
Texture Avg. &96.40 &97.19$\scriptstyle\pm0.3$ &99.44 &99.81$\scriptstyle\pm0.6$ &98.98 &98.84$\scriptstyle\pm0.3$ &96.64 &96.40$\scriptstyle\pm0.2$ &97.36 &97.62$\scriptstyle\pm0.1$\\
\midrule
Object Avg. &87.63 &90.37$\scriptstyle\pm0.1$ &97.53 &97.81$\scriptstyle\pm1.5$ &95.40 &97.30$\scriptstyle\pm0.4$ &84.03 &85.60$\scriptstyle\pm0.3$ &88.14 &89.10$\scriptstyle\pm0.2$\\
\midrule
\midrule
Overall Avg. &90.55 &\textbf{92.64}$\scriptstyle\pm0.2$ &98.17 &\textbf{98.48}$\scriptstyle\pm1.2$ &96.60 &\textbf{97.81}$\scriptstyle\pm0.2$ &88.23 &\textbf{89.20}$\scriptstyle\pm0.2$ &91.21 &\textbf{91.94}$\scriptstyle\pm0.3$\\
\midrule
\bottomrule
\end{tabular}
\caption{Comparison of TAB(Ours) and ImageNet pre-trained weights on SOTA methods for the image-level anomaly detection performance on MVTecAD\cite{bergmann2019mvtec} dataset. The abbreviation ‘ImgN.’ denotes ‘ImageNet'. The five random seeds of mean and standard deviation results are reported in AUROC(\%). Bold indicates the best performance. Please note that the asterisk (*) indicates using a pre-trained Industrial dataset to prevent overlap in the few-shot setting.}
\label{tab:detection}
\end{table*}

\subsection{Anomaly-Text-Aware Pre-training Strategy\label{Loss_OF}}

Since our text encoder $E_T$ is frozen, the output text features $f^{T}$ will be fixed. Thus, our training goal is to make the visual features $f^{V}$ extracted from our image encoder $E_V$ close enough to $f^{T}$ in the latent space for normal and abnormal data, respectively, as shown in Figure \ref{fig:optimized_feature}. 
We calculate the logits similarity matrix ${M}$ between $f^{T}$ and $f^{V}$ by applying Eq.(\ref{eq:eq4}) and (\ref{eq:eq5}). We optimize the distance between both features by using cross-entropy loss in the Eq.(\ref{eq:eq6}). Thus, the class name list $labels$, normal similarity matrix $M_n$, abnormal similarity matrix $M_a$, and final loss function ${L_{total}}$ will be described below. Assume input normal images be denoted by $X_n$, synthetic anomaly images be denoted by $X_a$, and its labels denoted as $y$, 

\begin{align}
\begin{split}
    X_n &= \{X_{1,n}, X_{2,n}, ..., X_{k,n}\},
    \\
    X_a &= \{X_{1,a}, X_{2,a}, ..., X_{k,a}\},
    \\
    y   &= \{y_1, y_2, ..., y_k\}.
\end{split}
\label{eq:eq3}
\end{align}
where $\hat{y}$ is the one hot encoding of $y$.




\begin{align}
\begin{split}
    f^{z}_s &= \{ E_z(X_s) | s\in\{n, a\} , z\in\{V, T\} \}
\end{split}
\label{eq:eq4}
\end{align}
where n, a, V, and T refer to the abbreviation of normal, abnormal, visual, and text, respectively. The shapes for both $f^{V}_n$ and $f^{V}_a$ are ${\mathbb{R}^{B*C}}$ and the shapes for both $f^{T}_n$ and $f^{T}_a$ are ${\mathbb{R}^{K*C}}$, where $B$ and $K$ refer to batch size and the number of classes in the pre-training data.

\begin{align}
\begin{split}
    M_s &= \{ f^V_s \cdot transpose(f^T_s) | s\in\{n, a\} \},
\end{split}
\label{eq:eq5}
\end{align}

\begin{align}
\begin{split}
    L_{CE}(q, p) &= -\sum_{x}p(x)\log q(x),
    \\
    L_{total} &= \frac{L_{CE}(M_n, \hat{y}) + L_{CE}(M_a, \hat{y})}{2}
\end{split}
\label{eq:eq6}
\end{align}
where $p(x)$ and $q(x)$ are ground truth and predicted output probabilities, respectively.




%% file: sec/expriments.tex
\section{Experiments}
\label{sec:experiments}

\subsection{Datasets}

To demonstrate the effectiveness of Anomaly Backbone, we conduct extensive experiments on some challenging real-world benchmark datasets, MVTecAD\cite{bergmann2019mvtec}, MixedWM38\cite{mixedwm38dataset} and implement on cross-dataset BTAD\cite{mishra2021vt}, KSDD2\cite{bovzivc2021mixed} to evaluate the generalization of our backbone weights. Each dataset is briefly described as follows:
\textbf{Industrial dataset\label{industrialdataset}} We propose an industrial dataset for manufacturing visual inspection. This dataset was constructed by selecting and collecting open-source datasets from various websites worldwide and official public conference workshops. The industrial dataset comprises 30 categories, including 15 object categories and 15 texture categories, totaling 17,393 images. All images in this dataset are normal. Image samples for each category are provided in the supplementary materials.

\textbf{MVTecAD} dataset\cite{bergmann2019mvtec} includes 5,354 images of ten industrial objects and five textures, with 3,629 normal and 1,725 test images, including 1,258 anomaly images. It provides image-level and pixel-level annotations.

\textbf{BTAD} dataset\cite{mishra2021vt} has 2,830 images in three classes of different resolutions. Each class includes normal and anomaly images, including scratch or distortion defect types.

\textbf{KSDD2} The KSDD2 dataset\cite{bovzivc2021mixed} developed with images of defective production items, contains 356 defective and 2,979 normal images.

\textbf{MixedWM38} dataset\cite{mixedwm38dataset}, a mixed-type wafermap defect dataset, includes real and synthetic patterns. It contains 1 normal pattern, 8 basic single defect patterns, and 29 mixed defect patterns. It provides a one-hot encoding defect label in 8 dimensions corresponding to the 8 basic defect types.


\begin{table*}
\centering
\renewcommand{\arraystretch}{0.5}
\tabcolsep=0.15cm
\begin{tabular}{c|cc|cc|cc|cc|cc}
\toprule
&\multicolumn{2}{c}{PaDiM\cite{defard2021padim}} &\multicolumn{2}{c}{RD4AD\cite{deng2022anomaly}} &\multicolumn{2}{c}{SimpleNet\cite{liu2023simplenet}} &\multicolumn{2}{c}{RegAD(4shots)\cite{huang2022registration}} &\multicolumn{2}{c}{RegAD(8shots)\cite{huang2022registration}}\\
\cmidrule(lr){2-3} \cmidrule(lr){4-5} \cmidrule(l){6-7} \cmidrule(l){8-9} \cmidrule(l){10-11}
Pre-trained &{ImgN.} &{TAB} &{ImgN.} &{TAB} &{ImgN.} &{TAB} &{ImgN.} &{TAB*} &{ImgN.} &{TAB*}\\
\midrule
Texture Avg. &95.42 &95.61$\scriptstyle\pm0.2$ &97.02 &97.53$\scriptstyle\pm0.8$ &92.05 &95.37$\scriptstyle\pm0.9$ &94.66 &94.90$\scriptstyle\pm0.5$ &95.26 &95.60$\scriptstyle\pm0.2$\\
\midrule
Object Avg. &97.06 &97.23$\scriptstyle\pm0.0$ &97.26 &98.45$\scriptstyle\pm0.9$ &95.94 &95.48$\scriptstyle\pm0.3$ &96.32 &96.73$\scriptstyle\pm0.6$ &97.38 &97.75$\scriptstyle\pm0.5$\\
\midrule
\midrule
Overall Avg. &96.51 &\textbf{96.69}$\scriptstyle\pm0.1$ &97.18 &\textbf{98.14}$\scriptstyle\pm0.8$ &94.64 &\textbf{95.44}$\scriptstyle\pm0.4$ &95.77 &\textbf{96.12}$\scriptstyle\pm0.5$ &96.67 &\textbf{97.03}$\scriptstyle\pm0.3$\\
\midrule
\bottomrule
\end{tabular}
\caption{Comparison of TAB(Ours) and ImageNet pre-trained weights on SOTA methods for the pixel-level anomaly localization performance on MVTecAD\cite{bergmann2019mvtec} dataset.  The abbreviation ‘ImgN.’ denotes ‘ImageNet'. The five random seeds of mean and standard deviation results are reported in AUROC(\%). Bold indicates the best performance. Please note that the asterisk (*) indicates using a pre-trained Industrial dataset to prevent overlap in the few-shot setting.}
\label{tab:localization}
\end{table*}


\subsection{Evaluation Metrics}
We utilize the standard Area Under the Receiver Operator Curve (AUROC) as our primary metric in the evaluation phase. The image-level AUROC is employed for anomaly detection, while the pixel-level AUROC is used for anomaly localization. The F1 score is employed for a fair comparison metric in the defect classification experiment.

\subsection{Experiments Settings}
In this study, we utilize the MVTecAD\cite{bergmann2019mvtec} and an Industrial dataset for pre-training with ResNet18 as our backbone structure. We only require normal images from all categories in the training set, ensuring no overlap with the test set. We enhance state-of-the-art methods \cite{defard2021padim, liu2023simplenet, deng2022anomaly, huang2022registration} by substituting their feature extractor weights with the weights from our pre-trained anomaly backbone. In the subsequent experiments, we also compare the performance of these methods using ImageNet pre-training. The terms TAB and TAB* denote our weights when the MVTecAD or Industrial dataset is utilized as the pre-training dataset. We ensure consistency by replicating the experimental settings and methods from their official repositories and adhering to each paper’s implementation approach for fair metric evaluation. Note that category-specific detail results are shown in the supplementary materials.

\subsection{Performance in Anomaly Detection}
Table \ref{tab:detection} presents the anomaly detection results of state-of-the-art (SOTA) methods on the MVTecAD. Our TAB enhances the average AUROC across all anomaly detection (AD) methods, maintaining consistent model parameters. Our pre-trained model enables PaDiM\cite{defard2021padim} and RD4AD\cite{deng2022anomaly} to achieve improved generalization, with accuracy enhancements in 13 out of 15 classes and an increase in average AUROC by 2.09\% and 0.31\%, respectively. For  SimpleNet\cite{liu2023simplenet}, it also achieves an overall AUROC improvement of 1.21\%, with a substantial 20.47\% enhancement in the ‘screw’ category.

\subsection{Performance in Anomaly Localization}
Anomaly localization results are presented in Table \ref{tab:localization}. Our TAB enhances anomaly localization accuracy across most categories in conjunction with three representative methods. PaDiM\cite{defard2021padim} and SimpleNet\cite{liu2023simplenet} achieve average AUROC of 96.69\% and 95.44\%, respectively, surpassing ImageNet pre-training results. RD4AD\cite{deng2022anomaly} records a significant improvement, achieving an average pixel AUROC of 98.14\% with a 0.96\% increase. 

\subsection{Performance in Defect Classification}
We further evaluate the robustness of our industrial inspection anomaly backbone through defect classification, as shown in Table \ref{tab:defectclassification}. This experiment incorporates a trainable linear layer atop the ResNet-18 backbone. Our TAB outperforms ImageNet, demonstrating superior accuracy and F1 scores, exceeding 0.1 and 0.03, respectively. Furthermore, with an unfrozen backbone, our TAB exhibits potential for fine-tuning, achieving a score of 0.98 in both metrics.

\subsection{Performance on Few-Shot Settings}
RegAD\cite{huang2022registration} is fine-tuned using 4 or 8 normal images per class for the few-shot setup. In scenarios with limited industrial data, our TAB* effectively enriches features, aiding RegAD in achieving 89.20\% and 91.94\% average image-level AUROC, with improvements of 0.97 and 0.73, respectively, as shown in Table \ref{tab:detection}. In Table \ref{tab:localization}, TAB* also yields improvements of 0.35 and 0.36 in localization. These experiments underscore the robustness of our weights, even with limited fine-tuning data.

\subsection{Visualization and Qualitative Results}
\begin{figure}
\begin{center}
\includegraphics[scale=0.18]{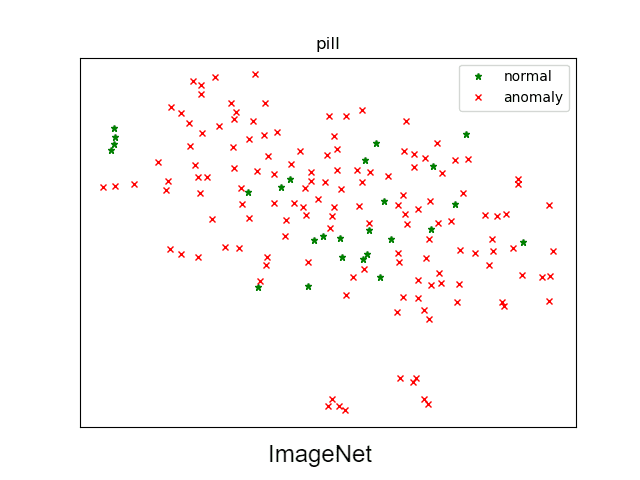}
\includegraphics[scale=0.18]{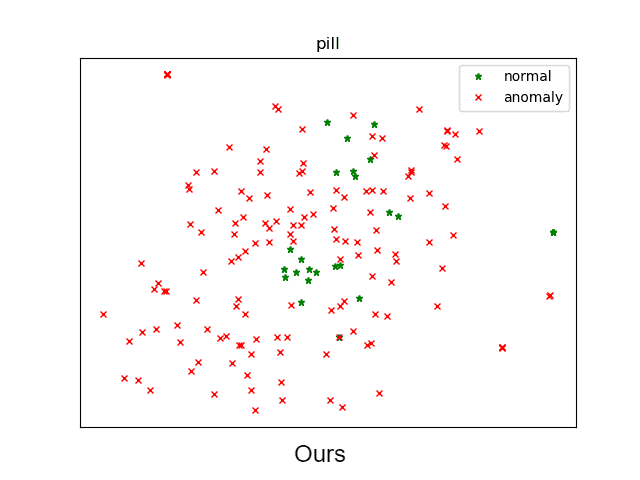}
\includegraphics[scale=0.18]{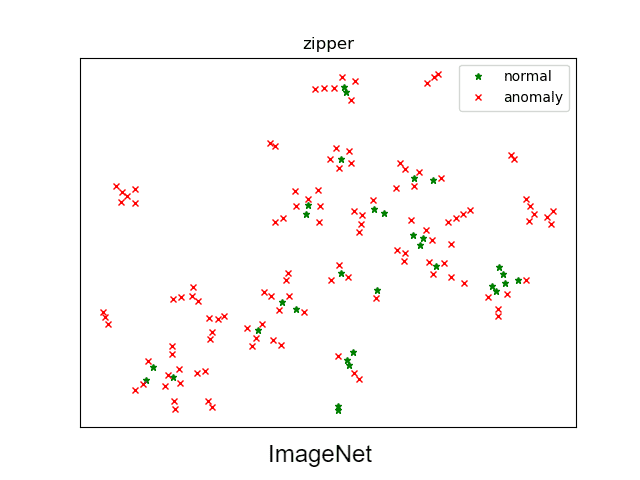}
\includegraphics[scale=0.18]{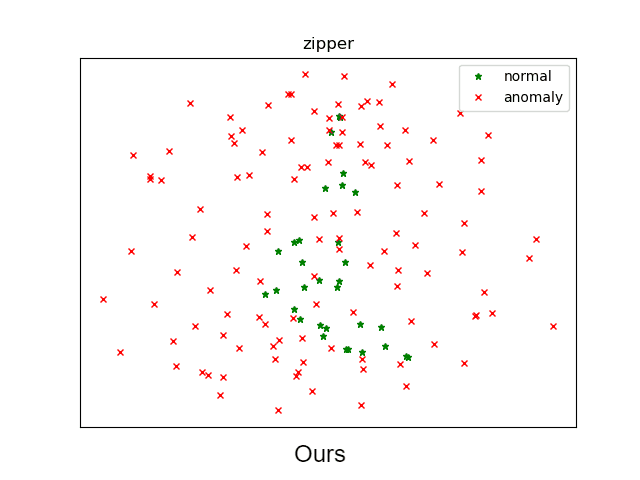}
\end{center}
   \caption{Visualization features of ImageNet and TAB(Ours) on Pill and Zipper categories. TAB(Ours) makes feature representation well-discriminate in normal features from anomaly features. Unlike ImageNet, our anomaly-text-aware pre-trained strategy distributes anomaly features around the normal feature cluster, providing a better initial decision boundary state.}
\label{fig:tsne}
\end{figure}

\begin{figure}
\begin{center}
\includegraphics[scale=0.10]{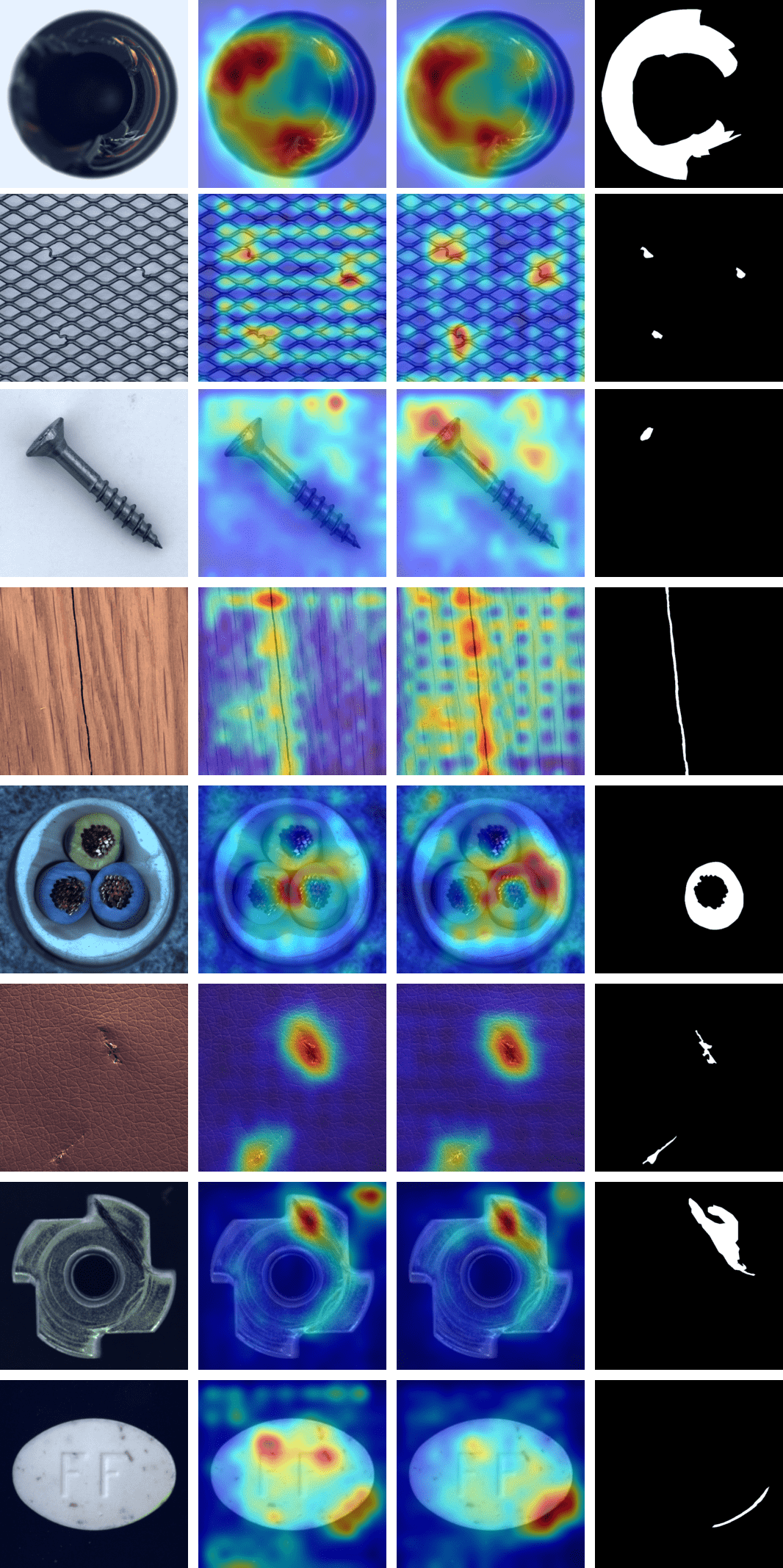}
\includegraphics[scale=0.10]{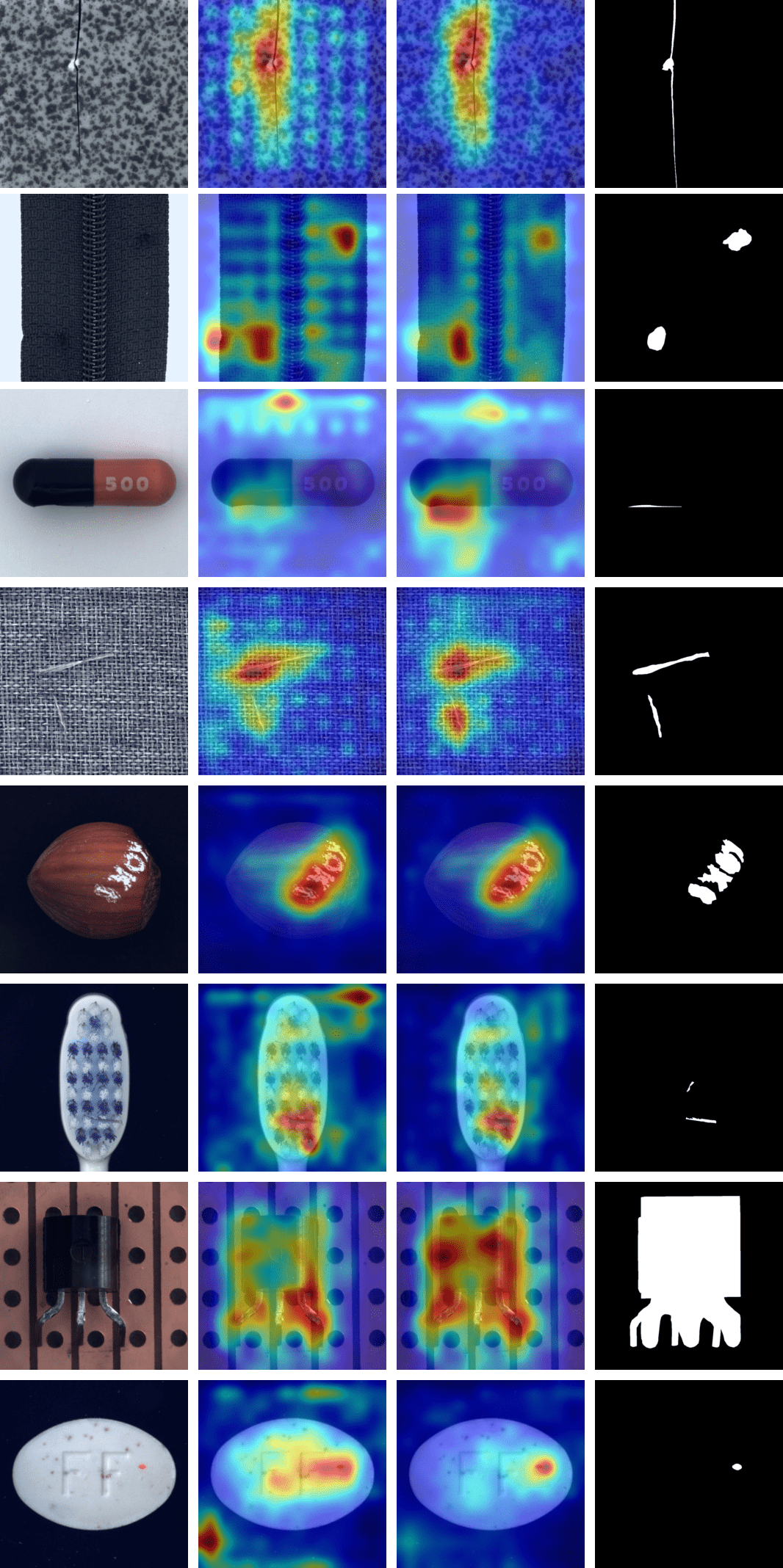}
\end{center}
   \caption{Visualization of RD4AD\cite{deng2022anomaly} localization results with different backbone weight. From left to right column, defect image, ImageNet, TAB(Ours), and ground truth.}
\label{fig:mvtecvisual}
\end{figure}

We adopt t-SNE\cite{van2008visualizing} for visualizing high-dimensional feature distributions. Here, we demonstrate the extracted features of ImageNet and ours in Figure \ref{fig:tsne}. As figure \ref{fig:tsne} shows, ImageNet represents the normal feature into a few clusters, and the initial decision boundary is implicit. In contrast, our anomaly-text-aware pre-trained strategy distributes anomaly features around a main normal features cluster to provide an explicit decision boundary. 
Figure \ref{fig:mvtecvisual} demonstrates the RD4AD method with anomaly localization heat map. The columns from left to right are defect image, ImageNet, Ours, and pixel-level ground truth, respectively. We can observe that ours can help RD4AD suppress background noise and output precise heat map results.

\subsection{Ablation Studies}

\subsubsection{Cross-dataset}


In this section, we assess the generalization of our anomaly backbone weights on novel categories to gauge overfitting. We utilize BTAD\cite{mishra2021vt} and KSDD2\cite{bovzivc2021mixed} as cross datasets (Table \ref{tab:crossdataset}). Our TAB outperforms ImageNet in detection and localization for all methods on the KSDD2 dataset. Additionally, it shows a substantial improvement in the mean AUROC over three BTAD classes. Despite PaDiM’s limited localization performance due to its training-free nature, we posit that a deeper trainable model could yield better results with our pre-trained weights.

\subsubsection{Effectiveness of Pre-trained Strategy}
We examine our anomaly-text-aware pre-training strategy alongside the ImageNet classification method under identical pre-training data settings to mitigate dataset bias. In Table \ref{table:pre-trained strategy detection}, ‘ImageNet’ refers to the reported results of each SOTA method, while ‘Classification’ denotes our approach of training the feature extractor backbone to classify images as normal or abnormal using cross-entropy loss, with training data comprising normal MVTecAD images and synthetic abnormal data. Our TAB demonstrates that our proposed anomaly-text-aware pre-training strategy, with the same pre-training data, is more effective in representing discriminative features by aligning visual and text features in both normal and abnormal contexts.

\subsubsection{Effectiveness of Synthetic Methods}
The authenticity and quality of synthetic abnormal samples significantly influence optimization performance. We provide an extensive analysis of synthetic methods across various industrial categories. As shown in Table \ref{tab:syntheticmethods}, each synthetic method is evaluated on PaDiM within our pre-training framework. NSA outperforms others in both detection and localization, as it generates synthetic data that accurately mirrors various industrial categories, thereby enriching our pre-training model. Figure \ref{fig:synthetic_samples} showcases examples of synthetic images and their corresponding ground truth.

\subsubsection{Effectiveness of Aligning Text Features}
Table \ref{tab:alignanomalytext} affirms the efficacy of incorporating both normal and abnormal text information in our framework. Our TAB exhibits lower detection performance than the Classification setting when aligned with only normal text. This is because it solely aligns normal images and text, whereas the Classification setting also considers abnormal images during pre-training, enhancing its detection performance. However, when aligned with normal and abnormal text, our TAB successfully discriminates between normal and abnormal visual features.

\subsubsection{Prompt Design}
Table \ref{tab:prompt_association} demonstrates that textual supervision for the term ‘anomalous’ enhances our framework by providing a contrasting concept of normal and abnormal during visual and text representation alignment (Two-Class). An integrated sentence prompt is crucial for infusing comprehensive semantic information into our backbone’s visual representation learning. Lastly, specific domain keywords (‘manufacturing’, ‘industrial’) offer a precise starting point for an enriched industrial image prompt.

\begin{table}[t]
\centering
\renewcommand{\arraystretch}{1.0}
\tabcolsep=0.13cm 
\begin{tabular}{c|c|cccc} 
\toprule    
    & Freeze & Accuracy &Precision & Recall &F1 \\ \hline   
    
ImageNet &V &0.22 &0.34 &0.28 &0.25 \\
TAB &V &\textbf{0.32} &\textbf{0.35} &\textbf{0.31} &\textbf{0.28}\\
\bottomrule
ImageNet & &0.95 &0.95 &0.95 &0.95 \\
TAB & &\textbf{0.98} &\textbf{0.98} &\textbf{0.98} &\textbf{0.98}\\

\bottomrule
\end{tabular}
\caption{Comparative analysis of ImageNet and TAB for the defect classification performance on MixedWM38\cite{mixedwm38dataset} dataset with the freeze and unfreeze backbone two protocols.}
\label{tab:defectclassification}
\end{table} 

\begin{table}[t]
\centering
\renewcommand{\arraystretch}{1.0}
\tabcolsep=0.09cm
\begin{tabular}{c|c|cc|cc} 
\toprule
    & &\multicolumn{2}{c}{Detection} & \multicolumn{2}{c}{Localization}\\ \hline   
    &Category & ImageNet  & TAB & ImageNet & TAB \\ \hline
\multirow{2}{*}{PaDiM} 
&KSDD2 & 68.90 & \textbf{70.30} & 94.90 & \textbf{95.50}\\ \cline{2-6}
&BTAD & 94.33 & \textbf{94.43} & \textbf{96.37} & 95.50\\ \hline
\multirow{2}{*}{SimpleNet} 
&KSDD2 & 70.24 & \textbf{70.47} & 79.32 & \textbf{88.04}\\ \cline{2-6}
&BTAD & 93.34 & \textbf{94.45} & 94.83 & \textbf{95.43} \\ \hline
\multirow{2}{*}{RD4AD} 
&KSDD2 & 92.50 & \textbf{95.60} & 97.30 & \textbf{97.60}\\ \cline{2-6}
&BTAD & 89.53 & \textbf{91.53} & 96.93 & \textbf{97.07} \\
\bottomrule
\end{tabular}
\caption{Cross-dataset experiment was set up with a pre-training phase on all normal samples from the MVTecAD dataset, followed by testing on the KSDD2 and BTAD datasets.}
\label{tab:crossdataset}
\end{table}
\begin{table}[ht]
\centering
\renewcommand{\arraystretch}{1.0}
\tabcolsep=0.13cm
\begin{tabular}{c|ccc|ccc}
\toprule
    &\multicolumn{3}{c}{Detection} &\multicolumn{3}{c}{Localization} \\ \cline{2-7}
    & ImgN. & Cls.  & TAB & ImgN. & Cls.  & TAB\\ \hline
    
PaDiM & 90.55 & 91.12 & \textbf{92.64} & 96.51 & 96.47 & \textbf{96.69}\\ \hline
SimpleNet & 96.60 & 96.96 & \textbf{97.81} & 94.64 & 94.45 & \textbf{95.44}\\ \hline
RD4AD & 98.17 & 98.31 & \textbf{98.48} & 97.18 & 97.13 & \textbf{98.14} \\ \hline
\end{tabular}
\caption{Impact of utilizing a pre-training strategy on detection and localization on the MVTecAD. The abbreviations ‘ImgN.’ and ‘Cls.’ denote ‘ImageNet’ and ‘Classification’, respectively.}
\label{table:pre-trained strategy detection}
\end{table}
\begin{table}[ht]
\centering
\renewcommand{\arraystretch}{1.0}
\tabcolsep=0.18cm 
\begin{tabular}{c|c|cc} 
\toprule
    &Synthetic Method & Detection & Localization\\ \hline   
\multirow{4}{*}{PaDiM} 
& Perlin & 90.79 & 96.26 \\
& Mask & 90.80 & 96.36 \\
& CutPaste & 91.52 & 96.42 \\
& NSA & \textbf{92.64} & \textbf{96.69} \\ \hline

\end{tabular}
\caption{Experiment with the effectiveness of including different synthetic methods in our proposed framework and evaluate the performance of PaDiM in the MVTecAD.}
\label{tab:syntheticmethods}
\end{table}
\begin{table}[ht]
\centering
\renewcommand{\arraystretch}{1.0}
\tabcolsep=0.19cm
\begin{tabular}{c cc cc} 
\toprule    
    &\multicolumn{2}{c}{Text-Align}  &\multicolumn{2}{c}{Task} \\ \cline{2-5}
    &Normal &Abnormal &Det. &Loc. \\ \hline

Classification & & &91.12 &96.47 \\
TAB(Ours) &V & &90.78 &96.54 \\
TAB(Ours) &V &V &\textbf{92.64} &\textbf{96.69} \\

\bottomrule
\end{tabular}
\caption{Experiment of evaluating the efficacy of text information alignment. PaDiM experimented on the MVTecAD. The abbreviations ‘Det.’ and ‘Loc.’ denote ‘detection’ and ‘localization’.}
\label{tab:alignanomalytext}
\end{table}
\begin{table}[ht]
\centering
\renewcommand{\arraystretch}{1.0}
\tabcolsep=0.30cm
\begin{tabular}{l|c|cc} 
\toprule
Method & Detection & Localization\\ \hline   
One-Class & 45.31 & 69.11 \\
Two-Class & 73.75 & 85.77 \\
+ State Ensemble & 89.74 & 95.73 \\
+ CLIP Template & 90.58 & 96.10 \\
+ Industrial Association & \textbf{92.64} & \textbf{96.69} \\ \hline

\end{tabular}
\caption{Experiment of the Industrial Prompt Association setting and evaluate the performance of PaDiM in the MVTecAD.}
\label{tab:prompt_association}
\end{table}

%% file: sec/conclusion.tex
\section{Conclusion}
\label{sec:conclusion}

In an effort to bridge the domain gap between the ImageNet dataset and the manufacturing dataset, we have introduced a novel Text-align Anomaly Backbone (TAB). This backbone, specifically pre-trained for industrial inspection tasks, employs an unsupervised approach. Our unique Industrial Domain Prompt Association (IDPA) and Anomaly-Aware-Text pre-training strategy significantly enhance the performance of our anomaly backbone, surpassing traditional ImageNet metrics in various industrial inspection downstream tasks.
Notably, we are the first to integrate anomaly text information into the industrial backbone pre-training paradigms. Our pre-trained weights have demonstrated robustness, consistently outperforming current state-of-the-art methods in cross-dataset and few-shot experiments.

%% file: main.bbl
\begin{thebibliography}{21}
\providecommand{\natexlab}[1]{#1}
\providecommand{\url}[1]{\texttt{#1}}
\expandafter\ifx\csname urlstyle\endcsname\relax
  \providecommand{\doi}[1]{doi: #1}\else
  \providecommand{\doi}{doi: \begingroup \urlstyle{rm}\Url}\fi

\bibitem[Batzner et~al.(2023)Batzner, Heckler, and K{\"o}nig]{batzner2023efficientad}
Kilian Batzner, Lars Heckler, and Rebecca K{\"o}nig.
\newblock Efficientad: Accurate visual anomaly detection at millisecond-level latencies.
\newblock \emph{arXiv preprint arXiv:2303.14535}, 2023.

\bibitem[Bergmann et~al.(2019)Bergmann, Fauser, Sattlegger, and Steger]{bergmann2019mvtec}
Paul Bergmann, Michael Fauser, David Sattlegger, and Carsten Steger.
\newblock Mvtec ad--a comprehensive real-world dataset for unsupervised anomaly detection.
\newblock In \emph{Proceedings of the IEEE/CVF conference on computer vision and pattern recognition}, pages 9592--9600, 2019.

\bibitem[Bo{\v{z}}i{\v{c}} et~al.(2021)Bo{\v{z}}i{\v{c}}, Tabernik, and Sko{\v{c}}aj]{bovzivc2021mixed}
Jakob Bo{\v{z}}i{\v{c}}, Domen Tabernik, and Danijel Sko{\v{c}}aj.
\newblock Mixed supervision for surface-defect detection: From weakly to fully supervised learning.
\newblock \emph{Computers in Industry}, 129:\penalty0 103459, 2021.

\bibitem[Defard et~al.(2021)Defard, Setkov, Loesch, and Audigier]{defard2021padim}
Thomas Defard, Aleksandr Setkov, Angelique Loesch, and Romaric Audigier.
\newblock Padim: a patch distribution modeling framework for anomaly detection and localization.
\newblock In \emph{International Conference on Pattern Recognition}, pages 475--489. Springer, 2021.

\bibitem[Deng and Li(2022)]{deng2022anomaly}
Hanqiu Deng and Xingyu Li.
\newblock Anomaly detection via reverse distillation from one-class embedding.
\newblock In \emph{Proceedings of the IEEE/CVF Conference on Computer Vision and Pattern Recognition}, pages 9737--9746, 2022.

\bibitem[Deng et~al.(2009)Deng, Dong, Socher, Li, Li, and Fei-Fei]{deng2009imagenet}
Jia Deng, Wei Dong, Richard Socher, Li-Jia Li, Kai Li, and Li Fei-Fei.
\newblock Imagenet: A large-scale hierarchical image database.
\newblock In \emph{2009 IEEE conference on computer vision and pattern recognition}, pages 248--255. Ieee, 2009.

\bibitem[Gudovskiy et~al.(2022)Gudovskiy, Ishizaka, and Kozuka]{gudovskiy2022cflow}
Denis Gudovskiy, Shun Ishizaka, and Kazuki Kozuka.
\newblock Cflow-ad: Real-time unsupervised anomaly detection with localization via conditional normalizing flows.
\newblock In \emph{Proceedings of the IEEE/CVF Winter Conference on Applications of Computer Vision}, pages 98--107, 2022.

\bibitem[Guo et~al.(2023)Guo, Jiang, Huang, Cheng, and Gong]{guo2023mldfr}
Yinghui Guo, Meng Jiang, Qianhong Huang, Yang Cheng, and Jun Gong.
\newblock Mldfr: A multilevel features restoration method based on damaged images for anomaly detection and localization.
\newblock \emph{IEEE Transactions on Industrial Informatics}, 2023.

\bibitem[Huang et~al.(2022)Huang, Guan, Jiang, Zhang, Spratling, and Wang]{huang2022registration}
Chaoqin Huang, Haoyan Guan, Aofan Jiang, Ya Zhang, Michael Spratling, and Yan-Feng Wang.
\newblock Registration based few-shot anomaly detection.
\newblock In \emph{European Conference on Computer Vision}, pages 303--319. Springer, 2022.

\bibitem[Jeong et~al.(2023)Jeong, Zou, Kim, Zhang, Ravichandran, and Dabeer]{jeong2023winclip}
Jongheon Jeong, Yang Zou, Taewan Kim, Dongqing Zhang, Avinash Ravichandran, and Onkar Dabeer.
\newblock Winclip: Zero-/few-shot anomaly classification and segmentation.
\newblock In \emph{Proceedings of the IEEE/CVF Conference on Computer Vision and Pattern Recognition}, pages 19606--19616, 2023.

\bibitem[Liu et~al.(2023)Liu, Zhou, Xu, and Wang]{liu2023simplenet}
Zhikang Liu, Yiming Zhou, Yuansheng Xu, and Zilei Wang.
\newblock Simplenet: A simple network for image anomaly detection and localization.
\newblock In \emph{Proceedings of the IEEE/CVF Conference on Computer Vision and Pattern Recognition}, pages 20402--20411, 2023.

\bibitem[Mishra et~al.(2021)Mishra, Verk, Fornasier, Piciarelli, and Foresti]{mishra2021vt}
Pankaj Mishra, Riccardo Verk, Daniele Fornasier, Claudio Piciarelli, and Gian~Luca Foresti.
\newblock Vt-adl: A vision transformer network for image anomaly detection and localization.
\newblock In \emph{2021 IEEE 30th International Symposium on Industrial Electronics (ISIE)}, pages 01--06. IEEE, 2021.

\bibitem[P{\'e}rez et~al.(2023)P{\'e}rez, Gangnet, and Blake]{perez2023poisson}
Patrick P{\'e}rez, Michel Gangnet, and Andrew Blake.
\newblock Poisson image editing.
\newblock In \emph{Seminal Graphics Papers: Pushing the Boundaries, Volume 2}, pages 577--582. 2023.

\bibitem[Radford et~al.(2021)Radford, Kim, Hallacy, Ramesh, Goh, Agarwal, Sastry, Askell, Mishkin, Clark, et~al.]{radford2021learning}
Alec Radford, Jong~Wook Kim, Chris Hallacy, Aditya Ramesh, Gabriel Goh, Sandhini Agarwal, Girish Sastry, Amanda Askell, Pamela Mishkin, Jack Clark, et~al.
\newblock Learning transferable visual models from natural language supervision.
\newblock In \emph{International conference on machine learning}, pages 8748--8763. PMLR, 2021.

\bibitem[Roth et~al.(2022)Roth, Pemula, Zepeda, Sch{\"o}lkopf, Brox, and Gehler]{roth2022towards}
Karsten Roth, Latha Pemula, Joaquin Zepeda, Bernhard Sch{\"o}lkopf, Thomas Brox, and Peter Gehler.
\newblock Towards total recall in industrial anomaly detection.
\newblock In \emph{Proceedings of the IEEE/CVF Conference on Computer Vision and Pattern Recognition}, pages 14318--14328, 2022.

\bibitem[Schl{\"u}ter et~al.(2022)Schl{\"u}ter, Tan, Hou, and Kainz]{schluter2022natural}
Hannah~M Schl{\"u}ter, Jeremy Tan, Benjamin Hou, and Bernhard Kainz.
\newblock Natural synthetic anomalies for self-supervised anomaly detection and localization.
\newblock In \emph{European Conference on Computer Vision}, pages 474--489. Springer, 2022.

\bibitem[Van~der Maaten and Hinton(2008)]{van2008visualizing}
Laurens Van~der Maaten and Geoffrey Hinton.
\newblock Visualizing data using t-sne.
\newblock \emph{Journal of machine learning research}, 9\penalty0 (11), 2008.

\bibitem[Wang et~al.(2020)Wang, Xu, Yang, Zhang, and Li]{mixedwm38dataset}
Junliang Wang, Chuqiao Xu, Zhengliang Yang, Jie Zhang, and Xiaoou Li.
\newblock Deformable convolutional networks for efficient mixed-type wafer defect pattern recognition.
\newblock \emph{IEEE Transactions on Semiconductor Manufacturing}, 33\penalty0 (4):\penalty0 587--596, 2020.

\bibitem[Xie et~al.(2023)Xie, Wang, Liu, Zheng, and Jin]{xie2023pushing}
Guoyang Xie, Jingbao Wang, Jiaqi Liu, Feng Zheng, and Yaochu Jin.
\newblock Pushing the limits of fewshot anomaly detection in industry vision: Graphcore.
\newblock \emph{arXiv preprint arXiv:2301.12082}, 2023.

\bibitem[Yang et~al.(2023)Yang, Soltani, and Darve]{yang2023anomaly}
Ziyi Yang, Iman Soltani, and Eric Darve.
\newblock Anomaly detection with domain adaptation.
\newblock In \emph{Proceedings of the IEEE/CVF Conference on Computer Vision and Pattern Recognition}, pages 2957--2966, 2023.

\bibitem[Yu et~al.(2021)Yu, Zheng, Wang, Li, Wu, Zhao, and Wu]{yu2021fastflow}
Jiawei Yu, Ye Zheng, Xiang Wang, Wei Li, Yushuang Wu, Rui Zhao, and Liwei Wu.
\newblock Fastflow: Unsupervised anomaly detection and localization via 2d normalizing flows.
\newblock \emph{arXiv preprint arXiv:2111.07677}, 2021.

\end{thebibliography}
